\documentclass[journal]{IEEEtran}
\usepackage{times}  
\usepackage{helvet} 
\usepackage{courier}  
\usepackage[hyphens]{url}  
\usepackage{graphicx} 
\usepackage{graphicx}  

\usepackage[normalem]{ulem}
\usepackage{xcolor}
\usepackage{xspace}
\usepackage{multirow}
\usepackage{booktabs}
\usepackage{amsmath,amssymb}
\usepackage{subfigure}
\usepackage{hyperref}

\newcommand{\name}[0]{d-SNEQ}
\newcommand{\sys}[0]{\name\xspace}
\newcommand{\sysq}[0]{\name$^\#$\xspace}
\newcommand{\eat}[1]{}

\newcommand{\mymarker}[1]{{  }}
\newcommand{\bm}[1]{\mathbf{#1}}

\ifCLASSINFOpdf
\else
\fi
\begin{document}
%
\title{Semi-supervised Network Embedding with Differentiable Deep Quantisation}
%
%
%

\author{Tao He,
        Lianli Gao, Jingkuan Song
        and Yuan-Fang Li
\thanks{T. He and Y. Li  are with the Faculty
of Information and  Technology, Monash University, Australia,
VIC, 3800, e-mail: tao.he@monash.edu and yuanfang.li@monash.edu. }
\thanks{L. Gao and J. Song are with the University of Electronic
Science and Technology of China, Chengdu, Sichuan 611731, e-mail: lianli.gao@uestc.edu.cn and jingkuan.song@gmail.com}
}

\maketitle

\begin{abstract}
Learning accurate low-dimensional embeddings for a network is a crucial task as it facilitates many downstream network analytics tasks. 
For large networks, the trained embeddings often require a significant amount of space to store, making storage and processing a challenge.  
  Building on our previous work on semi-supervised network embedding, we develop d-SNEQ, a differentiable DNN-based quantisation method for network embedding. d-SNEQ incorporates a rank loss to equip the learned quantisation codes with rich high-order information, and is able  to substantially compress the size of trained embeddings, thus reducing storage footprint and accelerating retrieval speed.   
We also propose a new evaluation metric, path prediction, to fairly and more directly evaluate model performance on the preservation of high-order information. 
Our evaluation on four real-world networks of diverse characteristics shows that \sys outperforms a number of state-of-the-art embedding methods in link prediction, path prediction, node classification and node recommendation, while being far more space- and time-efficient. 
\end{abstract}


\IEEEpeerreviewmaketitle

\section{Introduction}
\IEEEPARstart{N}etwork  embedding methods learn an encoder system that converts nodes into low-dimensional vectors, aiming at preserving the original network information as much as possible.  However, with the fast increase in network sizes, especially the number of nodes, it is increasingly challenging to store and retrieve such large-scale networks via the conventional embedding scheme, i.e.\ continuous embeddings, where each node is represented as a float-valued vector. Another issue of  continuous embedding methods  is that the learned embeddings usually contain massive  redundant information and many dimensions can be compressed~\cite{lian2018high,dblp:IPQ,DBLP:conf/sigir/ZhangZLG14}. 

The above two dilemmas motivate us to turn to more efficient embeddings, such as hashing and quantisation. In  the computer vision communities, these two strategies have been extensively studied for the task of image and video retrieval~\cite{wang2018survey,cao2016deep,DBLP:conf/aaai/SongHGXHS18}. The main advantages of these techniques are that the discrete representations can significantly reduce storage footprint and simultaneously improve retrieval efficiency. For storage footprint, the discrete vectors utilize a much smaller set of finite values. Specifically, hashing-based representations have only two possible values, zero and one, to denote each dimension, while quantisation has $\rm{K}$ different values for each dimension, also named as \emph{codewords} in some works~\cite{cao2016deep1,liu2018deep}, where $\rm{K}$  is usually greater than $2$. Thus, each dimension can be encoded by $\log _{2}^{K}$ bits which is much less than the storage size required by conventional embedding methods, e.g., $32$ bits or 64 bits for each dimension. 
For retrieval efficiency, due to the fact that each embedding dimension can only been selected from a finite set, when calculating the pointwise product and multiplication between two discrete vectors, the result of each dimension is also finite. Thus, we can pre-calculate all their possible finite results and store them in a fixed-size table with the size of $\rm{K}^2$. After that, the expensive online calculations of conventional methods can be sidestepped by efficient lookups in the pre-calculated table. 

Additionally, the other issue of network embedding is the usage of label information. Unsupervised node representation learning methods are the most common. These include the well-known methods DeepWalk~\cite{perozzi2014deepwalk}, LINE~\cite{tang2015line}, node2vec~\cite{grover2016node2vec}, Graph2Gauss (G2G)~\cite{DBLP:conf/iclr/BojchevskiG18}, and NetMF~\cite{qiu2018network}, all of which completely disregard the valuable node label information, leaving significant room for improvement on node classification performance. 
In fact, it is impractical to train a model in a fully supervised fashion, as we cannot expect the availability of ground-truth label for each node, especially for large networks. However, discarding the available labels is not helpful for the model to learn semantically discriminative representations. Naturally, semi-supervised or weakly supervised methods can well mitigate this problem, as they are able to make use of limited label information. Many semi-supervised network embedding methods have been proposed, including the recent techniques based on graph convolutional networks (GCN)~\cite{DBLP:conf/iclr/KipfW17,DBLP:conf/ijcai/Hu0WWT19}. GCN-based methods can be seamlessly assembled into a deep neural network in an end-to-end manner and facilitate many downstream applications, e.g.\ multi-label classification~\cite{chen2019multi}.  

A critical task to evaluate the quality of embeddings is \emph{link prediction}~\cite{perozzi2014deepwalk,tang2015line,DBLP:conf/iclr/KipfW17,dos2016multilabel}, which tests the capability of  embeddings to reconstruct the original network's neighbourhood structure, i.e., the first-order proximity. 
{On the other hand, it has been widely acknowledged that higher-order proximity has significant impact on the quality of embeddings~\cite{lian2018high,DBLP:conf/iclr/BojchevskiG18,zhu2018high,yang2017fast}.} {
	While some existing methods~\cite{DBLP:conf/iclr/BojchevskiG18,DBLP:conf/iclr/KipfW17,DBLP:conf/ijcai/Hu0WWT19} have claimed their strength in preserving high-order proximity, this capability is typically evaluated by indirect, alternative metrics, such as link prediction or node classification. Moreover, from our empirical observations, these tasks do not consistently reflect the preservation of high-order proximity.}
{ Moreover, Zhang et.\ al~\cite{zhang2018arbitrary} reported values of the objective function of  high-order proximities  preservation to measure it.} 
{However, those measurements are indirect too, as they do not empirically and explicitly demonstrate their ability to well embed proximity of higher orders. Therefore, a natural question arises, that 
	\begin{quote}
		\emph{Can we develop an explicit and unified metric for the evaluation of high-order proximity preservation beyond the indirect alternatives such as link prediction and node classification?}
\end{quote}}
{Toward this end, we propose a new metric, \emph{path prediction}.} {Intuitively, the preservation of high-order proximity gauges the model's capability of embedding hierarchical neighbourhood structure~\cite{du2018galaxy,ma2018hierarchical,chen2018harp}. Building on this intuition, we view path prediction as a multi-neighbourhood classification problem. That is, given a selected anchor node, we randomly sample its various high-order neighbours and train a classifier to predict those neighbours' order. We posit that if the high-order proximity is well-preserved in the embeddings, its neighbours  will be readily classified.} 
{Additionally, another practical application of path prediction is that we can directly obtain the multi-hop relationship between two nodes, e.g.\ second-hop neighbours, beyond simply answering whether there exists a direct edge between them by link prediction. Although we could construct the original network by link prediction an then use graph search algorithms, such as Depth First Search (DFS)~\cite{tarjan1972depth} or Breadth First Search (BFS) \cite{cormen2001introduction}, to identify paths of exact orders between two nodes, it is time-consuming and even infeasible for large-scale networks.
}

{In this paper, based on our preliminary work~\cite{he2020sneq}, we propose a semi-supervised network embedding method through differentiable deep quantisation. Specifically, the embedding component incorporates both node labels, network proximity information, as well as node attributes, and preserves high-order proximity. 
	Meanwhile, we deploy a direct differentiable quantisation module, built on a deep autoencoder neural network, to learn quantisation codes with a relaxation-free strategy. }
Our experiments on four real-world datasets with diverse characteristics show that our model outperforms existing state-of-the-art embedding methods on the tasks of link prediction, node classification, node recommendation and path prediction. 
Comparing to well-known discrete embedding methods, we also achieve substantially better performance on these tasks while maintaining a comparable storage footprint and time cost for retrieval. 

In summary, our main contributions are threefold: 

\begin{itemize}

	\item {Building on our preliminary work SNEQ~\cite{he2020sneq}, we  propose a differentiable DNN-based quantisation method, named d-SNEQ, which can provide quantised codes with more hierarchical high-order information by a rank loss. Compared with SNEQ, d-SNEQ can bring about  lower quantisation errors and improve its performance on the semantic and structural preservation by exploiting the merit of a deep autoencoder network. }
	
	\item We propose a new evaluation metric for higher-order proximity preservation, \emph{path prediction}, which more robustly measures the a model's performance in preserving high-order proximity.
	
	\item In our extensive experiments on link prediction, node classification, node recommendation and path prediction, our method outperforms state-of-the-art network embedding methods, including both continuous and discrete embeddings. 
\end{itemize} 

{Note that a preliminary version of this paper appeared in AAAI 2020~\cite{he2020sneq}. 
	In the preliminary version, we proposed a semi-supervised embedding scheme and a self-attention-based quantisation compression strategy. In this expanded manuscript, we concentrate on the quantisation module and make the following extensions:}
\begin{itemize}
	
	\item {We further extend the quantisation module in SNEQ to a deeper neural network version by carefully designing an autoencoder network to more effectively quantise embeddings.} 
	
	\item {We deploy a rank loss on the quantisation codes to force them to be equipped with consistent high-order proximities. } 
	
	\item  {To enable the quantisation module to be trained in an end-to-end manner, we utilise a more effective and general technique, Gumbel-softmax, which shows a better performance than our previous self-attention-based strategy. }
	
	\item {We propose a novel and direct evaluation metric, path prediction, for better evaluating a model's performance on higher-order proximity preservation. }
	
	\item {We thoroughly conduct a suite of ablation study experiments to test the contribution of each component. }
\end{itemize}

\section{Related Work}

Our work is closely related to learning  to quantisation and network embedding, so we briefly review the literature on those two tasks. 

\subsection{Learning to  Quantisation} 
Quantisation is a widely-used compression strategy to reduce amount of the redundant information in rich data, especially in the computer vision community~\cite{DBLP:journals/corr/abs-1902-00153,cao2017deep,DBLP:journals/pami/JegouDS11,DBLP:conf/aaai/CaoL0ZW16}. Generally, we can divide the mainstream high-dimensional data compression methods into two categories: binary hashing methods~\cite{wang2018survey} and   quantisation~\cite{DBLP:journals/pami/JegouDS11}. Specifically,   hashing methods  try to learn a hash function to map high-dimension data into a binary-valued space via multiple hyperplanes~\cite{erin2015deep,wang2018survey}, while the latter aims at quantising high-dimension data with a number of codewords, or centroids, so that the quantisation error is reduced as much as possible. Hence, in a wide range of empirical experiments, the performance of quantisation has been shown to be significantly better than hashing by a large margin. However, the main advantage of hashing is that it generally outperforms quantisation in terms of time and storage efficiency, because binary hashing only requires the storage of the indices of each data point, whilst quantisation needs to store both the indices and the centroids. In this work, we concentrate on quantisation and briefly review two quantisation techniques: Product Quantisation and DNN-based Quantisation. 

\textbf{Product Quantisation (PQ)} was proposed by J{\'{e}}gou~\cite{DBLP:journals/pami/JegouDS11} to quantise a vector into  several subvectors, each of which is represented by a selected codeword via vector quantisation. Subsequently, many subsequent works based on PQ have been proposed. An optimized product quantisation was proposed~\cite{ge2013optimized}, where authors introduced two optimization strategies: non-parametric and parametric methods, both of which can improve the  performance of approximate nearest neighbour (ANN) search. 
Norouzi and Fleet~\cite{norouzi2013cartesian} subsequently proposed a Cartesian K-means strategy with a compositional parameterization of cluster centroids that can boost the representational capacity of codewords and make it possible to quantise data using billions of codewords. Later on, non-orthogonal quantisations~\cite{babenko2014additive,chen2010approximate} was proposed, with the core idea of summing $M$ dependent codewords on the original space from $M$ different codebooks to approximate the raw vectors instead of concatenating $M$ subvectors. The benefit of summation is that   the compression quality can gain a significant improvement because $M$ can be an arbitrary size and not limited by the vector size, leading to little reconstruction  distortion. 

\textbf{DNN-based Quantisation} was investigated in some recent works~\cite{yu2018product,shu2017compressing,cao2017deep,morozov2019unsupervised}. Gao et al~\cite{cao2017deep} first proposed Deep Visual-Semantic Quantisation (DVSQ) to jointly learn deep visual-semantic features and visual-semantic quantisation codes in an end-to-end fashion, and it is able to significantly improve the quality of the learned compact codes and furthermore lift the similarity-based image retrieval performance. Shu and Nakayama ~\cite{shu2017compressing} developed a deep compositional code learning framework based on  multi-codebook quantisation (MCQ) in an end-to-end neural network to compress word embeddings in natural language processing and heavily reduce performance loss. Yu et al~\cite{yu2018product} developed a differentiable convolutional layer that is able to be plugged into an deep convolution feature learning network and made it possible to learn discrete codes in an end-to-end manner. Based on MCQ, Morozov and Babenko~\cite{morozov2019unsupervised} proposed an unsupervised visual representation compression method which can be generally applied in a wide range of computer vision pipelines for  compressed-domain retrieval. 

\subsection{Network Embedding}
Network embedding has been an active research area in representation learning~\cite{cui2018survey}. We can broadly divide network embedding methods into two categories: unsupervised and semi-supervised. The unsupervised methods include DeepWalk~\cite{perozzi2014deepwalk}, Line~\cite{tang2015line}, NetMF~\cite{qiu2018network}, and those based on autoencoders, such as VGAE~\cite{kipf2016variational},  and ARVGA~\cite{pan2018adversarially}. Specifically, DeepWalk~\cite{perozzi2014deepwalk} first proposed to use a sequence-to-sequence model to learn structural embeddings via a two-step strategy. Tang et al developed the influential two-step method LINE~\cite{tang2015line}, which is able to simultaneously learn the first- and second-order proximity. Later on, NetMF \cite{qiu2018network} theoretically unifies these methods under a matrix factorization framework. On the autoencoder-based unsupervised methods, VGAE~\cite{kipf2016variational} first utilised an autoencoder framework to reconstruct the adjacency matrix, and subsequently ARVGA~\cite{pan2018adversarially} devised a Variational Autoencoder (VAE) network that embeds each node representation into a Gaussian distribution and used Kullback–Leibler (KL) divergence to force the reconstructed distributions to be consistent with the original ones. 

Semi-supervised network embeddings has also gained extensive attention. Even through label information may be scarce, it is of potential value for the model to learn discriminative features, especially with the invention of graph convolution networks (GCN)~\cite{yang2019triplet,liang2018semi,hamilton2017inductive}. TAE~\cite{yang2019triplet} is an end-to-end model-free framework based on metric learning and can not only preserve the topological structure of a network but also embed the discriminative semantic signals. SEANO~\cite{liang2018semi} is a semi-supervised method and  able to preserve three types of critical information: attribute affinity, topological proximity and label similarity of nodes. GraphSAGE~\cite{hamilton2017inductive}  based on GCN can inductively embed network structure by sampling and aggregating features from a node's neighbours.

\section{Methodology}

In this section, we will present our network embedding and quantisation technique in detail. The overall architecture is shown in Fig.~\ref{fig.cifar}.

\textbf{Problem definition.}
Let ${G}=(V,{X},E,Y)$ represent an attributed network. 
$V=\{v_1,v_2,\ldots, v_N\}$ is a set of $N$ nodes. 
${X} \in {{\mathbb{R}} ^{N \times D}}$ represents $D$-dimensional node attributes. 
$E=\{e_{i,j}\}_{i,j=1}^N$ denotes the edge set of the network. 
Specifically, $e_{i,j}=1$ if $v_i$ has a connection to $v_j$, otherwise $e_{i,j}=0$. 
$Y=\{y_1,y_2,\ldots, y_N\}$ is the label set of each node. 
To simplify the notation, we use an adjacency matrix ${A} \in  \{0,1\}^{N\times N}$    to represent the edges of the network. 
Our embedding goal is to embed the nodes of the network $G$ into low-dimensional vectors $\bm{Z} \in {\mathbb{R}^{N \times L}}$, where $L \ll D$. 
Simultaneously, the quantisation module is responsible for encoding the real-valued vectors $\bm{Z}$ into short and compact codes $\bm{Q} \in \{0,1,\ldots,K\} ^{N \times M}$  and a set of codebooks $\mathcal{C} =\{{\bm{C}}_1,{\bm{C}}_2,\ldots,{\bm{C}}_M\}$, where ${\bm{C}}_i \in {\mathbb{R}^{K \times L}}$. 
$M$ is the arbitrary size of codebooks in the embedding space $\bm{Z}$, and  $K$ is the number of codewords in each codebook ${\bm{C}}_i$ and usually set to the power of $2$.

\subsection{Network Structure Embedding via Adaptive Margin}\label{sec:margin}

\begin{figure*}[ht]
	\centering
	\includegraphics[width=0.98\linewidth]{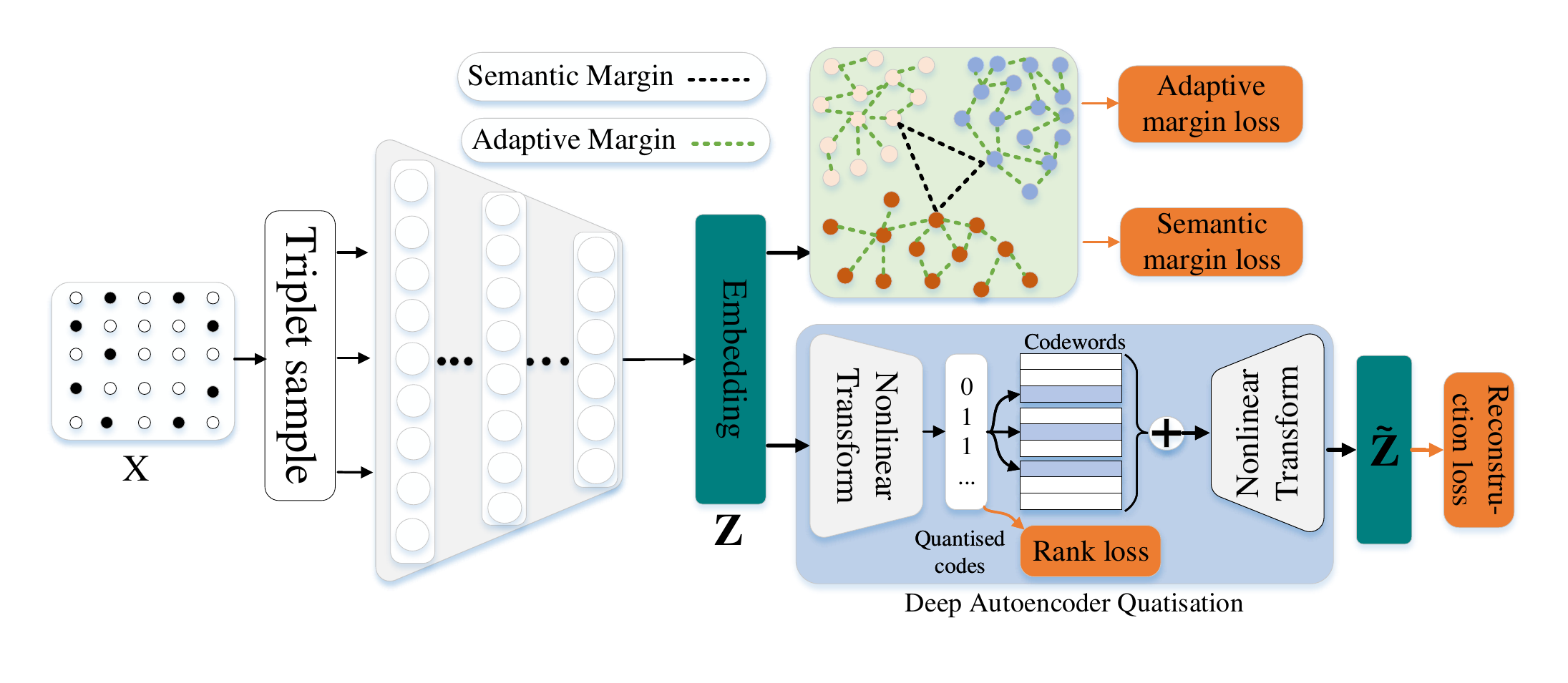}
	\caption{The high-level architecture of our framework. 
		\sys consists of three main parts: (1) network structure preservation through adaptive margin learning, (2) semi-supervised semantic margin learning, and (3) DNN-based quantisation learning. 
		The adaptive margin loss aims at forcing the embedding space to preserve the original neighbourhood information. 
		The semantic margin loss is responsible for separating nodes with different labels. 
		The   quantisation loss guarantees the reconstructed embeddings close to the original one. 
	}
	\label{fig.cifar}
\end{figure*}

The most important property of embeddings is the preservation of precise neighbourhood structure information of the original networks. To this end, we propose an adaptive margin based on metric learning as shown in \ref{fig.cifar}.
For an attributed network $G$, our model takes the attributes ${X}$ as the input. 
We use a multi-layer perceptron (MLP) to embed each node $v_i$, initialised with node attributes $\textbf{x}_i$: 
\begin{equation}
\bm{h}_j =  
f(\bm{w}_j{\textbf{x}_i} + \bm{b}_j)
\end{equation}
where $j$ denotes the $j$-th hidden layer and $f(.)$ is the activation function. 
For notational convenience, the output of the last layer is denoted by $\bm{Z}$. 


Intuitively, in a homogeneous network, the shorter the distance between two nodes is, the more similar they are. 
In this paper, we propose to use the shortest distance between two nodes (in terms of hops) to measure their structural similarity.
Theoretically, we can regard embedding as a procedure to project the shortest distance information in the original network into the embedding space. 
Our motivation is how to precisely preserve the different shortest distance into the embeddings so that  different distances of embeddings can simulate  hierarchical information in the network. 
To this end, we adopt the widely-used metric learning technique, which has been successfully applied in computer vision tasks such as person re-identification~\cite{DBLP:journals/ijon/DongLZLJG18}, face recognition~\cite{DBLP:conf/cvpr/SchroffKP15}, and hashing-based image retrieval~\cite{DBLP:conf/aaai/SongHGXHS18}, 
and  leverage the triplet loss to learn node distances. 
The adaptive margin loss $\ell_a$ is defined as below: 
\begin{align}
\begin{split}
{\ell_a} = \frac{1}{N} \mathop {\min} \limits_{\bm{W}^e} ( \sum\limits_{tr(a,p,n)}^N \max (&D{(\bm{z}_{a},\bm{z}_p)} - {D {(\bm{z}_a,\bm{z}_n)}} + \\[-8pt]
& {\delta _{a,n}} - {\delta _{a,p}},0) ) 
\label{eq.am}
\end{split}
\end{align}
where $\bm{W}^e$ denotes the parameters of the embedding network; $D(.)$ is the distance function in the embedding space, for which we chose Euclidean distance ($L_2$ norm); and $\delta_{i,j}$ denotes the shortest distance from node $v_i$ to node $v_j$. 
$tr(.)$ is a triplet sampling function that, given an anchor node $v_a$, selects a positive node $v_p$ and a negative node $v_n$. 
Concretely, given an anchor node $v_a$, $v_p$ and $v_n$ are sampled under the condition $\delta_{a,p} < \delta_{a,n}$. 

In Eq.~\ref{eq.am}, the {adaptive margin}, ${\delta _{a,n}} - {\delta _{a,p}}$, is the key term that defines the disparity between two shortest distances, that is, the anchor node to the positive node and the negative node. We adopt an adaptive margin to model the difference between a node's neighbours. Specifically, the more distant neighbour is divided by a large margin while the closer neighbour is separated by a smaller distance. 
There are many different algorithms to calculate the shortest distance $\delta$, such as Dijkstra's and Floyd's algorithms. In this paper we adopt a fast strategy based on matrix multiplication~\cite{sankowski2005shortest}. 

\subsection{Semi-supervised Semantic Embedding}\label{sec:semi}
In many situations, a subset of the nodes in a network are assigned labels, and intuitively, nodes with the same label should be closer in the embedding space than otherwise. 
Based on this intuition, we explore a semi-supervised strategy to preserve label information into embeddings $\bm{Z}$. 
The goal is to cluster same-label nodes with a small margin but separate those with different labels by a large margin. 
The general idea is illustrated in Figure~\ref{fig.cifar}, where the differently labelled nodes (different colors) should be separated by large margins. 

A key optimisation of our semi-supervised learning procedure is that we do not need to use all labelled nodes to train the model.
This is due to the hypothesis that nodes with the small shortest distance  are more likely to belong to the same class. 
Generally, as long as a small part of nodes in the same class are adequately separated, other nodes with small shortest distances to those separated nodes can also be separated.  
Thus, we only need to ensure that a  margin, name as semantic margin used to separate differently labelled embeddings, is large enough because if it is too small, the adaptive margin will conflict with it and degrade node classification performance. Furthermore, provided that the semantic margin is large enough, we can deem that it divides the embedding space into several discriminative subspaces, each of which stands for a community with the same class label, in which the adaptive margin is applied to preserve their neighbourhood structure information. 
The formulation of our semantic margin loss $\ell_c$ is given as the following:
\begin{equation}  
{\ell_c} = \frac{1}{T}  
\mathop {\min } \limits_{\bm{W}^e}
{\sum\limits_{i,j=1}^T {({{D(i,j)} - {S_{i,j}}} )}^2}
\label{eq.cm}
\end{equation}
where $T$ is the sample size for semi-supervision; $D(.)$ is the same distance function as in Eq.~(\ref{eq.am}); and $S_{i,j}$ represents the constant semantic margin and is formulated as below:
\begin{equation}
{S_{i,j}} = \left\{ {\begin{array}{*{20}{l}}
	{0,}&{{\rm{if}}\;{y_i} \cap {y_j} \ne \emptyset }\\
	{{M_c},}&{{\rm{otherwise}}}
	\end{array}} \right.
\end{equation}
where $y_i$ is the set of labels of node $v_i$, and $M_c$ is a constant defining the semantic margin. 

The aim of Eq.~(\ref{eq.cm}) is to force the same-labelled nodes to be close to each other, but nodes with different labels to be at a distance of $M_c$. 
It is worth noting that the value of $T$ is important to node classification performance, as a large $T$ value generally improves node classification result, due to more supervised information made available to the model. 
At the same time, a larger $T$ could negatively affect graph structure learning due to the fact that some nodes which are supposed to be neighbours are separated by a large margin of $M_c$, leading to a degradation of the structural preservation.
Hence, the choice of $T$ is a trade-off between two competing factors: structural and semantic information. In the experiments, we test the impact of different $T$.

\subsection{DNN-based Differentiable Quantisation}
As we mentioned previously, our quantisation strategy aims at compacting the continuous  embedding vectors. 
Ideally, this reduction would improve both space and time efficiency while preserving task performance. 
Product quantisation (PQ)~\cite{DBLP:journals/pami/JegouDS11} is a well-known quantisation technique. 
PQ quantises embedding vectors $\textbf{Z}$ into $M$ codebooks $\mathcal{C} =\{\bm{C}_1,\bm{C}_2,\ldots,\bm{C}_M\}$, and each codebook $\bm{C}_j\in\mathbb{R}^{{K\times L}}$ consists of $K$ centres (codewords) in the embedding space, which can be seen as $K$ subspaces in the embedding space. 
For each codebook $\bm{C}_{j}$, we use a 1-of-$K$ indicator vector $\bm{q}_{ij} \in \{0,1\}^{1\times K}$ to assign one centre to $\bm{z}_i$. 
It is worth noting that there is only one 1 in each $\bm{q}_{ij}$, which represents the closest centre to $\bm{z}_i$, while all the other values of $\bm{q}_{ij}$ are 0.  
The general  quantisation loss $\ell_q$ is calculated by:
\begin{equation}
{\ell_q} = \frac{1}{N}
\mathop {\min } \limits_{ \bm{W}^e,\bm{C}}
{\sum\limits_{i = 1}^N {\left\| {{\bm{z}_i} - \sum\limits_{j = 1}^M {{\bm{q}_{ij}}{\bm{C}_j}} } \right\|} ^2}
\label{eq.pq1}
\end{equation}

This loss function aims at reducing the reconstruction loss between the embedding vector $\bm{z}_i$. 
A smaller $\ell_q$ means better reconstruction performance from the quantisation codes. However, in our preliminary study~\cite{he2020sneq}, we observe that simply deploying $M$ codebooks cannot adequately reduce the reconstruction errors. Inspired by some recently proposed bottleneck networks~\cite{DBLP:conf/aaai/SongHGXHS18,shen2020auto}, we deploy a deep autoencoder-based network  to quantise $\bm{z}_i$ into  compact codes $\bm{\tilde{{z}}}_i$, where the bottleneck layer is the learned quantised codes and codewords, as shown in Figure~\ref{fig.cifar}. Generally, the enhanced quantisation loss can be rewritten as:
\begin{equation}
{\ell_q} = \frac{1}{N}
\mathop {\min } \limits_{\bm{W}^e,\bm{W}^{de},\bm{W}^{en},\bm{C}}
{\sum\limits_{i = 1}^N {\left\| {{\bm{z}_i} -  \mathcal{D}( \sum\limits_{j = 1}^M {{\mathcal{E}(\bm{z}_i)}_j{\bm{C}_j}}  )  } \right\|} ^2}
\label{eq.pq}
\end{equation}
where     $\bm{W}^{en}$ and $\bm{W}^{de}$ are the parameters of the encoder $\mathcal{E}$ and the decoder of $\mathcal{D}$, respectively. 
Noting that the output of  $\mathcal{E}(\bm{z}_i)$ is $M$ binary vectors, denoting the selected index of  codewords in the referring $M$ codebooks, that is $\bm{q}^i_{j} = \mathcal{E}(\bm{z}_i)_j$. 

Nevertheless, the optimisation of Eq.~(\ref{eq.pq}) becomes a discrete code search problem, that is, we need to find $M$ binary codes $\bm{q}$ to  approximate $\bm{z}$ as much as possible, similar to the learning to hash problem~\cite{wang2018survey}. 
The direct optimisation is an NP-hard problem. Thus,  many  works~\cite{LiWK16,DBLP:conf/sigir/ZhangZLG14}  turn to a relaxation strategy from hard (discrete) to  soft (continuous).  
Similarly, we can rewrite Eq.~(\ref{eq.pq}) in a continuous form:
\begin{align}
\setlength{\abovedisplayskip}{1pt}
{{\ell _q} = }&{\frac{1}{N}\mathop {\min }\limits_{\bm{W}^e,\bm{W}^{de},\bm{W}^{en},\bm{C}} \sum\limits_{i = 1}^N {{{\left\| {\bm{z}_i - \mathcal{D}( \sum\limits_{j = 1}^M {\bm{u}_{ij}\bm{C}_j}) } \right\|}^2}}  \text{ such that }}  \label{eq.pq_continuous}\\[4mm]
{}&{\bm{u}_{ij}=\mathcal{E}(\bm{z}_i)_j  \propto \bm{q}_{ij},\forall j = 1,2, \ldots ,N{\rm{ }}} \nonumber \\[1mm]
{}&{\bm{u}_{ij}\in {{[0,1]}^K}, {{\left\|\bm{u}_{ij} \right\|}_1} = 1, \forall j = 1,2, \ldots ,N{\rm{ }}} \nonumber \\[1mm]
{}&{\bm{q}_{ij} \in {{\{ 0,1\} }^K},{{\left\|\bm{q}_{ij} \right\|}_1} = 1,\forall j = 1,2, \ldots ,N{\rm{ }}} \nonumber
\end{align}

The next question is how to choose a relaxation strategy.  
A dominant strategy in hash learning is to use $sigmoid$ or $tanh$ activation functions to anneal the relaxed vectors $\mathbf{x}$ to $(0,1)$ and then force them to be close to $sign(\mathbf{x})$. A drawback of this approach is that it is hard to train the network  when the relaxed vectors are heavily distributed in the margin of $sigmoiod$ or $tanh$ function, where the gradient is close to zero, leading to large quantisation errors.
Inspired by the wide application of Gumbel-softmax~\cite{shu2017compressing,zhang2019doc2hash,maddison2016concrete}, we  deploy a differentiable discrete layer based on Gumbel-softmax so that the model can directly optimize quantisation codes in the training stage. Specifically, Gumbel-softmax can be presented as: 
\begin{align} 
\bm{u}_{ij} = softmax(\frac{\log(\mathcal{E}(\bm{z}_i)_j+g_i)}{\tau}) 
\label{eq:gumb}
\end{align}
where  $g_i = -\log (-\log (\mathrm{Uniform}(0,1)))$ denotes a standard Gumbel distribution and $\tau$ is a  temperature factor as  defined in ~\cite{maddison2016concrete}.
To obtain the discrete codes, we  choose the  maximum value in $\bm{u}_{ij}$ as the selected codewords and then convert its index into a one-hot vector, which can be defined as the below:
\begin{align}
\label{eq.one_hot}
\bm{q}_i &= \left[~ \bm{one\_hot}(\rm{argmax}(\bm{u}_{ij})) ~\right]_{j=1,\ldots, M}  
\end{align}
where $\bm{one\_hot(.)}$ is a function to convert the index to a one-hot vector; $\rm{argmax(.)}$ selects the index of the maximum value in a vector; $\left[. \right]$ denotes the concatenation operation.  

Though we add the adaptive margin on the continuous embedding $\bm{z}$, to furthermore boost the quantisation codes to be equipped with the neighbourhood proximity, we also deploy a rank loss over  quantisation codes, as the below:
\begin{align}
\begin{split}
{\ell_r} =   \mathop {\min} \limits_{\bm{W}^e,\bm{W}^{en},\bm{C}}\sum\limits_{tr(a,p,n)} \max (  \bm{u}_a \cdot  \bm{q}_n -\bm{u}_a \cdot\bm{q}_p +  
1,0) 
\label{eq.am1}
\end{split}
\end{align}
where $\cdot$ denotes the inner product of two vectors, $tr(a,p,n)$ is the sample function same as in Equation~(\ref{eq.am}), $\bm{u}$ is the probability output by Gumbel-softmax in Equation~(\ref{eq.pq_continuous}), and $\bm{q}_p$ is an one-hot vector in Equation~(\ref{eq.one_hot}). By optimising Equation (\ref{eq.am1}), it will force the index of the maximum  in $\bm{u}_a$ to be the same with $\bm{q}_p$, that is, letting the value of the index be close to $1$ but the others decrease to zero. On the other hand, for the negative points, it will reduce the value of $\bm{u}_a$ in the index of the maximum of $\bm{q}_n$ and increase the other dimensions' values to force the index of the maximum in $\bm{u}_a$  to differ from $\bm{q}_n$. Thus, the quantisation codes $\bm{q}$ is also equipped with the consistent structural information with the continuous embeddings.

\subsection{Learning}\label{sec:opt}
Our model  learns network embedding and quantisation in an end-to-end architecture, which integrates adaptive margin loss (Eq.~\ref{eq.am}), semi-supervised semantic margin loss (Eq.~\ref{eq.cm}),   reconstructed quantisation loss (Eq.~\ref{eq.pq_continuous}) and further a rank loss over discrete quantisation codes (Eq.~\ref{eq.am1}) into a joint optimisation problem:
\begin{equation}
\label{eq.joint}
\ell = \mathop {\min } \limits_{\bm{{W}^e,{W}^{en},{W}^{de},\bm{C}}} (\ell_a+\ell_r + \alpha \ell_c + \beta \ell_q)
\end{equation}
where $\alpha$ and  $\beta$ are the balance parameters to trade-off the importance of each part. 
All the parameters in Eq.~(\ref{eq.joint}) are differentiable and trained by mini-batch stochastic gradient descent. 

\section{Experiments}\label{sec:exp}
\noindent{\textbf{Datasets.}} We evaluate our method on four real-world networks. 
Brief statistics of the datasets are shown in Table~\ref{tab.datasets1}. It is worth noting that cDBLP~\cite{yang2015defining} is a large-scale   network.
We conduct experiments on the standard tasks of link prediction, node classification and node recommendation. Besides, we evaluate the model  on our newly proposed task, \emph{path prediction}. Finally, we evaluate the efficiency and effectiveness of quantisation. 
\eat{An ablation study, included in the supplementary material, is conducted to examine the contribution of each component of \sys.}
\begin{table}[!h]
	\centering
	\caption{Brief statistics of the datasets.}
	\label{tab.datasets1}
	\begin{tabular}{l*{5}{r}}
		\toprule
		$G$  & Cora\_ML   & Citeseer &      DBLP     & cDBLP \\ \midrule
		$|V|$    & 2,995 & 4,230         &  17,716 & 317,080       \\ 
		$|E|$    & 8,416 & 5,358     &  105,734   &   1,049,866 \\ 
		Attr.    & Yes             & Yes            & Yes     &  No \\ 
		Labels   & 7     & 6                  & 6  & 5,000\\ 
		\bottomrule
	\end{tabular}
\end{table}

\noindent\textbf{Implementation Details.} 
Our embedding method is a multi-layer deep neural network consisting of three dense layers, each of which is followed  by a  batch-norm layer and activated by ReLu. The encoder network of $\mathcal{E}$ consists of two dense layers, also followed by a batch-norm layer and a ReLu layer, and the decoder $\mathcal{D}$ is symmetrical with the encoder.
The dimension of the network embeddings is set to $128$. 
The initial learning rate $\eta$ is set as $0.001$ and we adopt the One Cycle learning rate schedule~\cite{smith2019super} for faster model convergence, and batch size is set to $100$.  
Hyperparameters $\alpha$ and $\beta$  are not fixed but tuned in an unsupervised way similar to \cite{xie2018learning}.  
Specifically, $\alpha =\frac{0.1}{{1 + {{\mathop{\rm e}\nolimits} ^{ - (\omega \mu) }}}} $, where $\omega$ is a constant value  $0.5$,  and $\mu$ is the training progress from $0$ to $1$, while $\beta$ is set as $1.0 - \frac{1}{{1 + {{\mathop{\rm e}\nolimits} ^{ - (\omega \mu) }}}}$. The temperature $\tau$ is set to $1$. 

For quantisation, the number of codebooks $M$ is  set to $8$  and the size of codebook $K$ is set to $256$, so each quantisation code takes up $64=(8*\log^{256}_2)$ bits. 
The amount of labelled nodes $T$ used in semi-supervised training (as defined in Eq.~(\ref{eq.cm}) is set to 10\% of $|V|$ by default. For a fair evalutaion, the compared semi-supervised methods also use the same ratio of labelled nodes for training. 
For a fair comparison with learning to hash methods, their hash dimension is set as $64$ bits, equal to our quantisation code length. 
All experiments were performed on a workstation with  $256$ GB memory, 32 Intel(R) Xeon(R) CPUs (E5-2620 v4 @ 2.10GHz) and 8 GeForce GTX 1080Ti GPUs. 
For each model a maximum approx.\ 100 GB of memory was allocated. 

\subsection{Baselines}
To demonstrate the effectiveness and efficiency of our method, we select the following representative and state-of-the-art network embedding methods as our comparison baselines. The baseline models consist of both continuous and discrete embedding methods.

\noindent\textbf{Continuous network embedding models.} We compare our continuous embedding model, denoted \sys, with the following methods:
\begin{itemize}
	\item DeepWalk (DW)~\cite{perozzi2014deepwalk} was the first to utilise a sequence-to-sequence model  in natural language processing to learn neighbour structure proximity via a two-step strategy. 
	
	\item TADW~\cite{yang2015network} is based on DeepWalk and it proposed to use rich information as auxiliary features to improve the quality of network representation by factorising a referring matrix.
	
	\item SEANO~\cite{liang2018semi} is a semi-supervised embedding method for attributed networks and can simultaneously embed three kinds of information: the topological proximity, label similarity of vertices, and attribute affinity.
	
	\item Graph2Gauss (G2G)~\cite{DBLP:conf/iclr/BojchevskiG18} was the first to propose to embed a node as a Gaussian distribution, which is represented as a mean vector and a variance vector.
	
	\item H-GCN~\cite{DBLP:conf/ijcai/Hu0WWT19}, based on graph convolutional networks, developed a graph pooling mechanism by coarsening nodes into a hyper node to enlarge the receptive field of the GCN, which enbles it to capture more global information. 
	\item TAE~\cite{yang2019triplet} is based on metric learning and it  utilises a triplet loss to learn the topological structure and preserve more discriminative information.
	
	\item ONE~\cite{bandyopadhyay2019outlier} focuses on attributed networks and is an unsupervised, outlier-aware network embedding method, by minimising the effect of outlier nodes to	improve the quality of the embeddings.
\end{itemize}

\noindent\textbf{Discrete network embedding models.} We compare our discrete embedding model, denoted \sysq, with the following methods:

\begin{itemize}
	\item  SH~\cite{datar2004locality} is a classical and   widely-applied learning to hash method for the approximate nearest neighbour search task.
	
	\item Discrete Collaborative Filtering (DCF)~\cite{zhang2016discrete}  is a principled hashing method able to tackle the challenging discrete optimization problem in hash learning and avoid large quantisation errors caused by two-step optimization.
	
	\item NetHash~\cite{wu2018efficient} utilises the randomized hashing technique to embed trees in a graph, which can preserve information closer to the root node as much as possible.
	
	\item Binarized Attributed Network Embedding (BANE)~\cite{yang2018binarized} develops  a  Weisfeiler-Lehman proximity matrix
	that can preserve the dependence between node attributes and connections via combining the features from neighbouring nodes. 
	
\end{itemize}

\subsection{Link Prediction}

Link prediction is a standard task to evaluate the performance of network embedding methods, aiming at measuring the preservation of the first-order  proximity. 
Specifically, we randomly select 5\% and 10\% edges as the validation and test set respectively, similar to Graph2Gauss~\cite{DBLP:conf/iclr/BojchevskiG18}. 
Following the convention, we use AUC as the performance metric. 

Table~\ref{tab.linkpre} summaries the link prediction results of the continuous embedding methods (upper block) and of the discrete methods (lower block). 
N/A denotes the algorithms that did not finish within 24 hours or within 100GB memory. 
From the upper block of Table~\ref{tab.linkpre}, it can be observed that our method consistently outperforms other continuous network embedding methods, except lower than G2G by $0.41$ percentages  on the Cora\_ML dataset. However, on the large dataset cDBLP, ours exceeds G2G by $3.96$ percentages. 
{Note that our model obtains very close performance with SNEQ~\cite{he2020sneq} in term of continuous embeddings. This is due to the fact that our modification focuses on improving quantisation codes instead of network embeddings. Hence, we do not report the results of d-SNEQ on continuous embeddings on the other evaluation tasks.}  
In the lower block of the table, it can be seen that our quantisation embedding is substantially superior to all the others, with at least 3$\sim$4 percentage points better than the second best method BANE. 
The possible reason is that as the embedding dimension is relatively low, hash-based methods suffer larger loss on structure information without an additional schema (codebooks). 
In contrast, with the same dimension, the codewords of our quantisation are able to preserve more structural information. 
We can also observe another advantage of our method, that our model can handle large-scale networks such as cDBLP while some hashing methods, including DCF and BANE, cannot. This is because all of them depend on SVD, which incurs high memory usage. 

\begin{table}[htp]
	\caption{Link prediction results of the continuous methods (top) and discrete methods (bottom). Best results in each block are \textbf{bolded}.}
	\centering
	\label{tab.linkpre}
	\begin{tabular}{lccccc}
		\toprule
		{Models} & Citeseer & {Cora\_ML} & DBLP  & cDBLP \\ \midrule
		DW       & 82.14    & {81.26}   & 71.67 & 66.84 \\
		TADW       & 85.58    & {84.40}   & 78.81 & 71.32 \\
		
		SEANO    & 81.61    & {84.42}   & 89.43 & 82.13 \\ 
		H-GCN   & 91.62    & {90.72}   & 88.57 & 81.47 \\ 
		TEA      & 92.70    & {92.81}   & 91.49 & 88.45 \\
		ONE    & 93.21    & {91.03}   & 93.72 & N/A     \\ 
		G2G     & 96.39  & \textbf{97.63}   & 97.71 & 88.18 \\
		SNEQ  &  {96.47}    & {97.21}   & \textbf{97.74} & {92.12} \\  
		\textbf{\sys}& \textbf{96.50}    & {97.22}   &   {97.72} &   \textbf{92.14} \\ \midrule
		SH         & 80.41    & {84.50}   & 83.31 & 80.57 \\ 
		DCF     & 79.42    & {80.67}   & 84.12 & N/A    \\ 
		NetHash   & 84.34    & {86.34}   & 88.51 & {83.72}    \\ 
		BANE      & 90.29    & {91.65}   & 90.43 & N/A    \\ 
		SNEQ$^\#$    &  {92.72}    &  {93.18}   &  {93.60} & {91.24}   \\ 
		\textbf{\sysq}      & \textbf{93.84}    & \textbf{94.24}   & \textbf{94.71} & \textbf{93.54}   \\ \midrule
	\end{tabular}
\end{table}

\subsection{Node Classification}
\begin{figure*}[!h]
	\centering
	\subfigure[\small{$F_1$ score on Citeseer}]{
		\includegraphics[width=0.32\linewidth,  ]{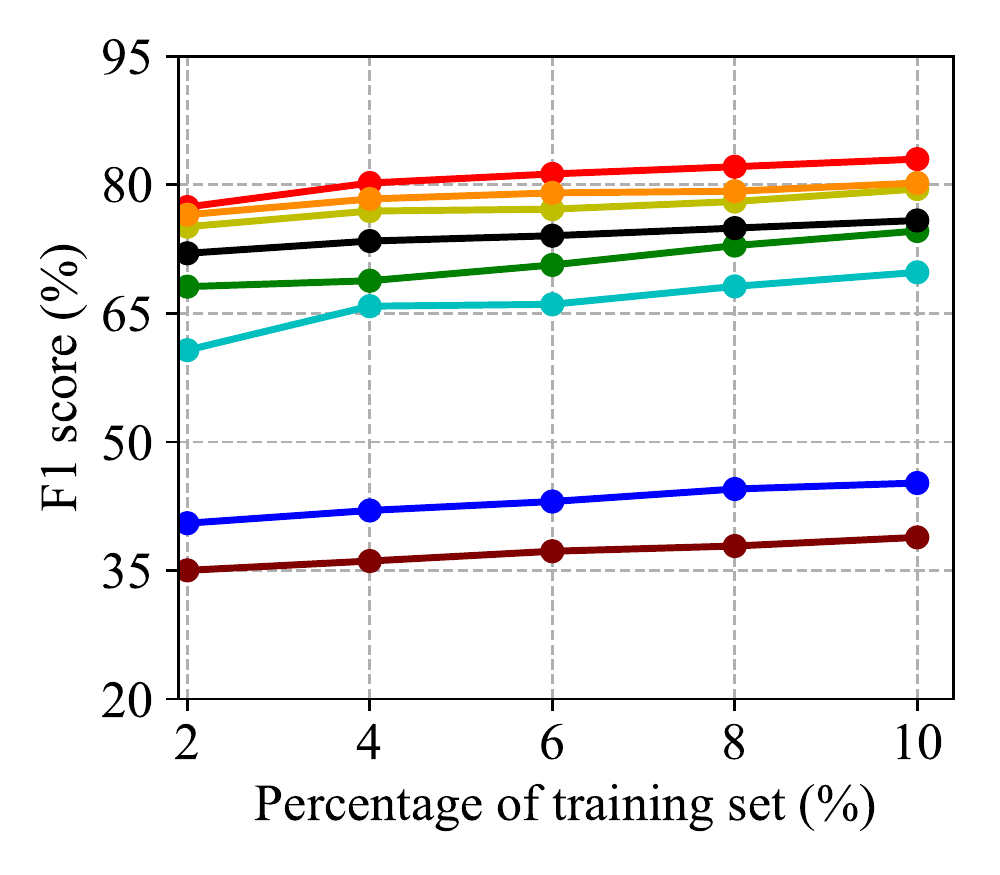}}
	\label{fig.a}
	\subfigure[\small{$F_1$ score on Cora\_ML.} ]{
		\includegraphics[width=0.32\linewidth,  ]{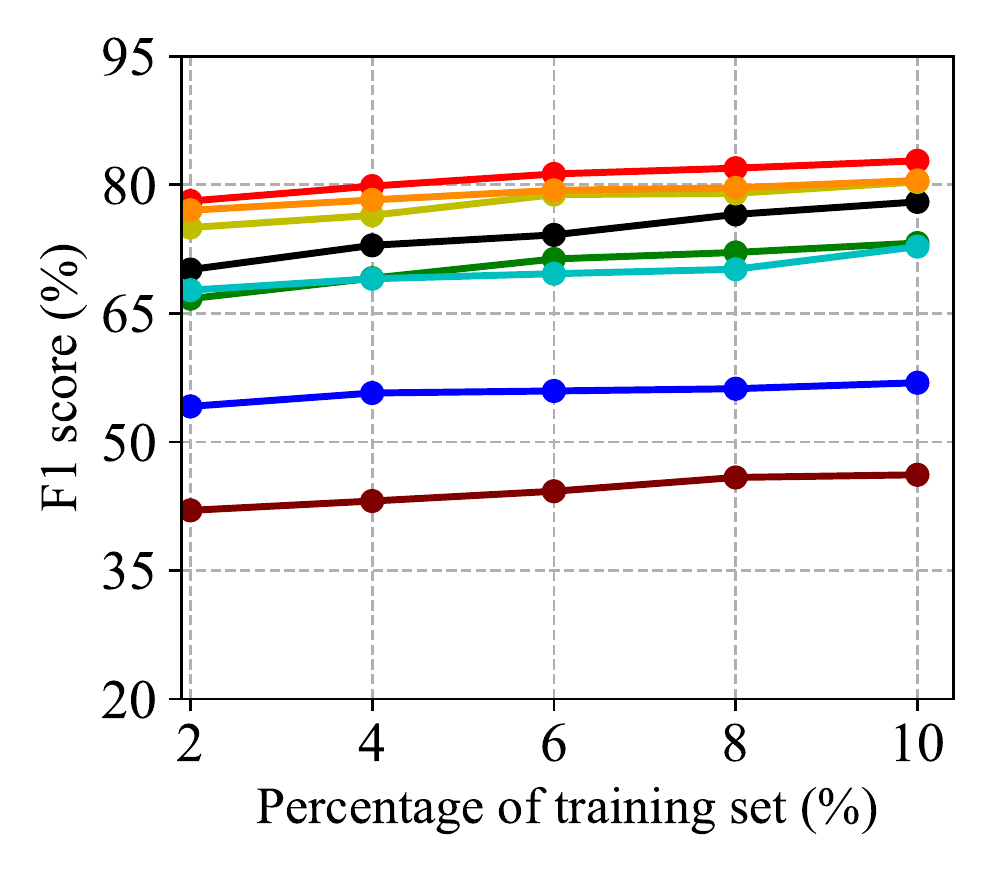}}
	\label{fig.b}
	\subfigure[ \small{$F_1$ score on DBLP.}]{
		\includegraphics[width=0.32\linewidth, ]{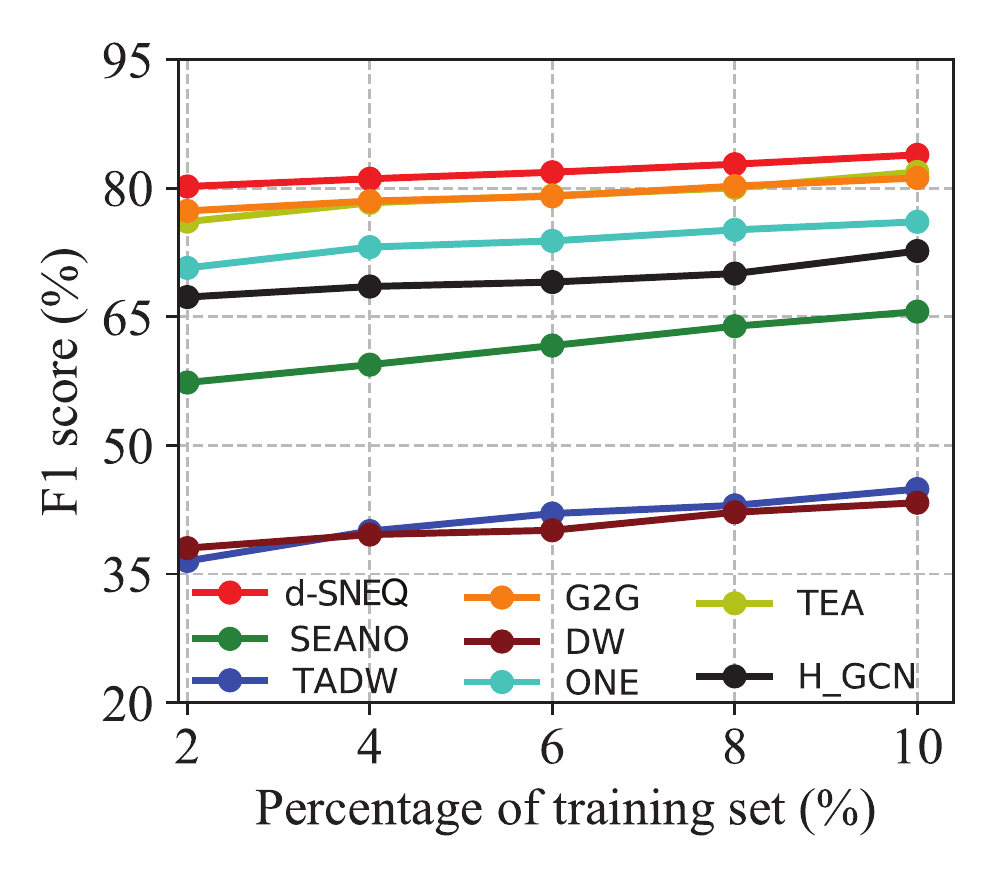}}
	\label{fig.c}
	
	\subfigure[\small{$F_1$ score on Citeseer.}]{
		\label{fig.d}
		\includegraphics[width=0.32\linewidth, ]{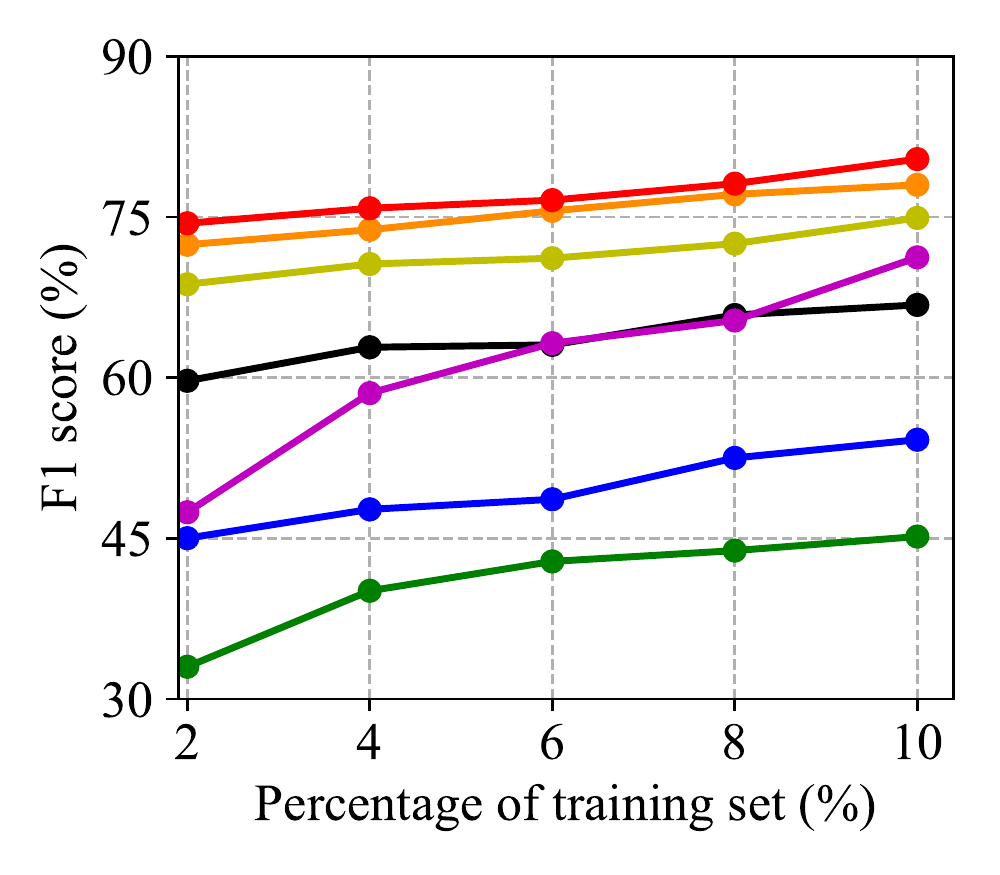}}
	\subfigure[\small{$F_1$ score on Cora\_ML.}]{
		\label{fig.e}
		\includegraphics[width=0.32\linewidth, ]{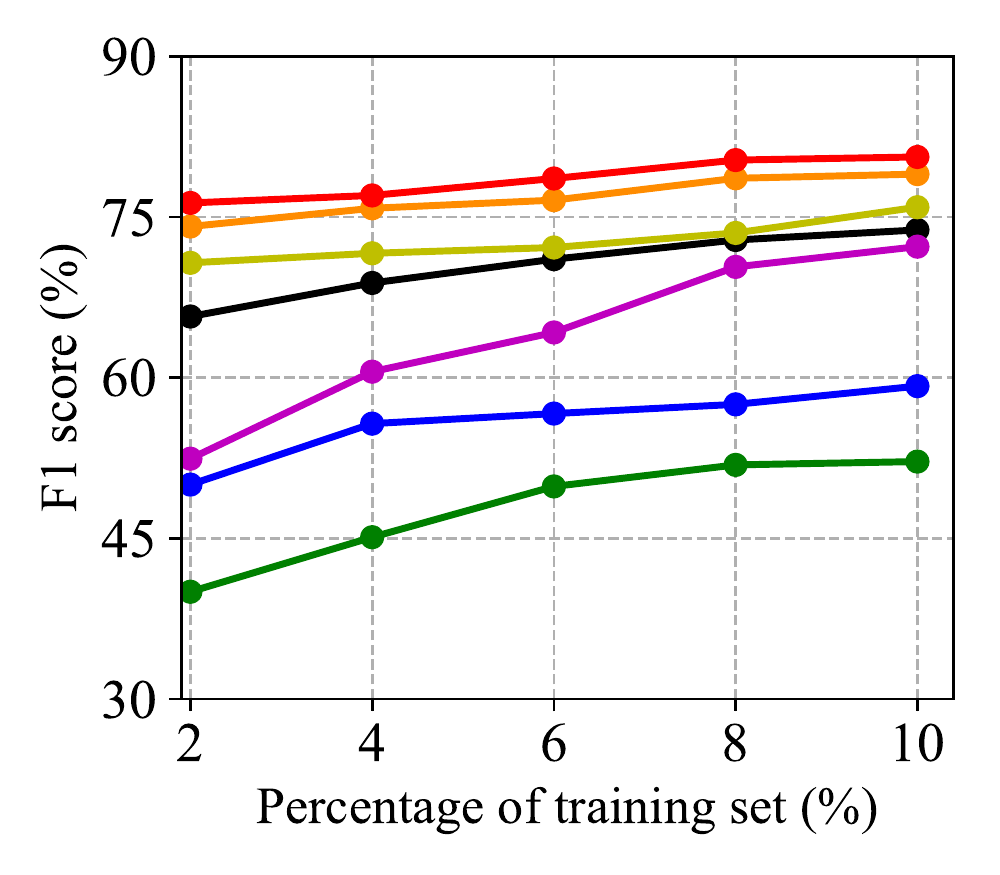}}
	\subfigure[\small{$F_1$ score on DBLP.}]{
		\label{fig.f}
		\includegraphics[width=0.32\linewidth, ]{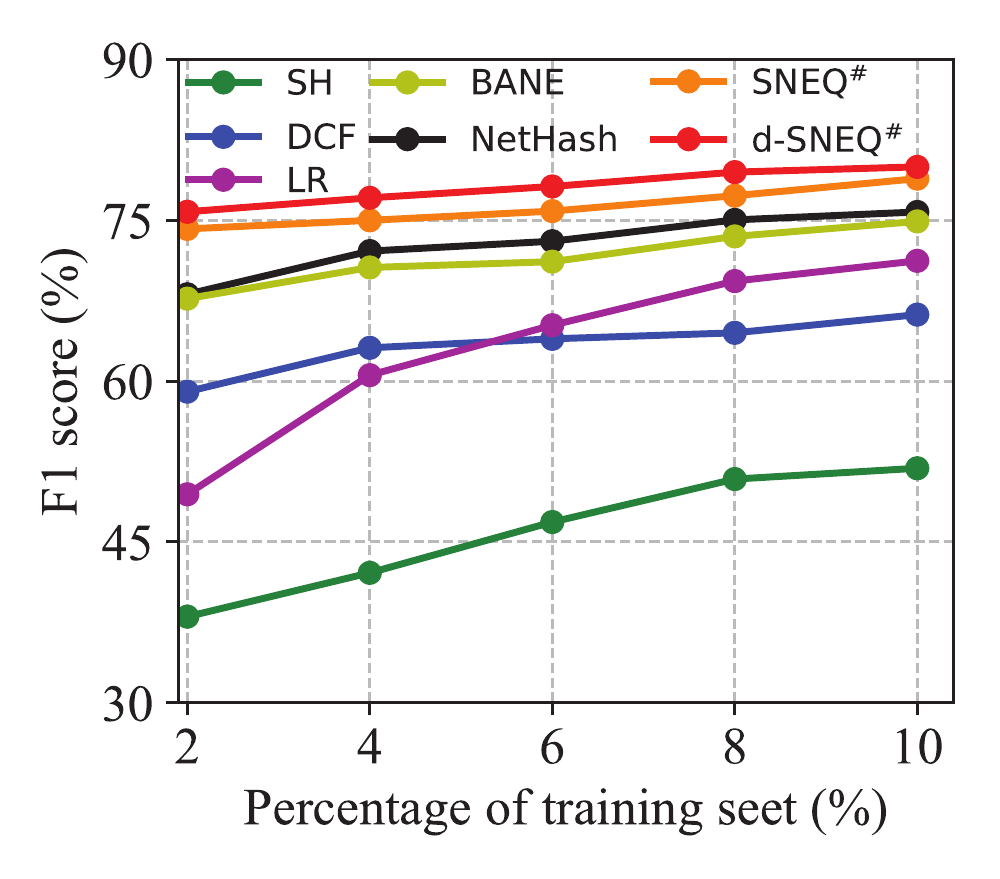}}
	\caption{Node classification results on the three attributed networks. The top row shows the results of continuous embeddings while the bottom row shows the results of discrete embedding results.}
	\label{fig.node_class}
\end{figure*}

We test our method on the three attributed datasets: Citeseer, Cora\_ML, and DBLP. 
We use the one-vs-the-rest logistic regression as the classifier, repeat the prediction for $10$ times, and report the average of Macro-F$_1$ and Micro-F$_1$ results. 
Figure~\ref{fig.node_class} shows the results, where Figures~\ref{fig.node_class} (a)--(c) are the results of the continuous embedding methods while (d)--(f) present the discrete results. 
In each experiment, a varying percentage (e.g.\ 2, 4, 6, 8, 10\%) of labelled nodes are sampled for training the classifier. 

From Figure~\ref{fig.node_class}, we can make the following observations.

(1) From Figures~\ref{fig.node_class}(a)--(c), we can observe that our continuous embeddings are superior to the counterparts on all of the three datasets. 
Specifically, our method is about 1$\sim$2 percentage points higher than the second best method G2G. 
H-GCN generally outperforms ONE since H-GCN performs a graph coarsening operation, grouping the nodes with same neighbours into one node. 
However, the node refining operation in H-GCN highly depends on the initial network structure and the complexity or sparsity of networks. 
In contrast, our semantic margin module does not depend on the neighbourhood structure, and only considers the pairwise distance between different nodes, and thus can generalise better for more complex networks.

(2) In the discrete embedding results, our quantisation shows a great advantage compared with the hashing-based methods. 
Specifically, our model exceeds the best hashing method BANE by about 4$\sim$5 percentage points on all of the three datasets. 
The possible reason is that the short binary codes lead to large loss of information in hashing methods. 
Meanwhile, as node attributes are binary and can be regarded as hash codes, we also test the classification performance of the raw attributes via a logistic regression classifier, denoted as LR. 
It can be observed that LR exceeds many hash methods such as NetHash and DCF. This is because the dimensions reduce significantly from $\sim3,000$ to 64 bits and many meaningful features are abandoned. 

It is worth noting that SH is trained by our continuous embeddings. However it suffers a huge performance degradation, compared from the continuous embeddings, of nearly 30 percentage points. 
We conjecture that there are two main reasons: (1) the dimensions reduce to about one third, leading to significant information loss; and (2) the procedure of learning to hash is not an end-to-end procedure, i.e., first learning embedding vectors and then training the hash function. 
In contrast, our quantisation avoid these issues in two ways: (1) the codewords can preserve as much information as possible, even though the quantised code length are short; and (2) the learning to embedding and   quantisation steps are unified in one network and jointly trained by the back propagation scheme.

\subsection{Path Prediction}
Though many methods incorporate the ability to handle higher-order proximity, they are not {explicitly} evaluated with a metric that is specifically designed to measure the preservation of higher-order proximity. 
Hence, we propose an extended link prediction task, dubbed \emph{path prediction}, that is designed to not only measure whether there exists a  connection between two nodes, but also to recognise which order it is between the two nodes, e.g., first-order, seconder-order or even higher-orders. Formally, given two nodes, the \emph{path} between them is defined as the \emph{minimum} hop length that connects them. For instance, if two nodes are immediate neighbours, the number of hops between them is $1$ and their path is also $1$. If two nodes are not immediate neighbours but are indirectly connected via another node, their path is $2$. By that analogy, we can define any two nodes' path by a adjacent matrix, denoted as $\bm{P} \in \mathbb{R}^{N \times N}$. Note that if two nodes are disconnected, their path is defined as \emph{infinite}. For evaluation purposes, we treat different path lengths as different categories and use the embeddings to train a classifier for path prediction. 

Towards this goal, we treat the path prediction as a multi-class classification problem.  
Theoretically, the more high-order proximity the embeddings are equipped with, the better performance the classifier gains under the same training configuration.  In terms of   experimental settings, we set the high-order range from the first to the fourth, that is the classifier is required to classify five classes, including the none class if two nodes do not have a path or their path length is greater than 5.    Inspired by the relation recognition task in scene graph generation~\cite{zellers2018neural,gu2019scene}, we use the multiplication of two embeddings, $\bm{v}_i * \bm{v}_j$, as the fusion feature to train a one-vs-the-rest logistic regression classifier under different training ratios of labelled points: $20\%$, $40\%$, $60\%$, and $80\%$. In addition, we randomly select $1,000$ pairs of nodes for each type of paths and report the average of Macro-F$_1$ and Micro-F$_1$ results. 

Table \ref{tab.path_pred} shows the comparison results of path prediction on the three attributed networks. Generally, we can obtain the following observations:

(1) Our method  consistently outperforms the other embedding models. Though G2G is able to preserve high-order information, ours is consistently better than it by about 2$\sim$3 percentages in the three datasets. We deem that it is because G2G only employs a fixed margin for graph structure preservation, leading to its inability to preserve more discriminative order information. By contrast, our model can learn different metrics for neighbours at different distances. 

(2) Furthermore, DW and TADW can also embed high-order proximity. However, both of them utilise a two-step strategy and thus are less effective than end-to-end models based on deep neural networks such as G2G and TEA. Similarly,  \sysq exceeds the best discrete model BANE by approx.\ 3$\sim$4 percentages, which demonstrates the advantage of quantisation over hash methods. 

{(3) Although some methods, such as ONE and NetHash, perform well on the link prediction and node classification tasks, their performance deteriorates substantially on the path prediction task. This result demonstrates that link prediction, the widely-used evaluation metric, cannot consistently reflect a model's capability to preserve high-order proximity. For example, while ONE achieves competitive performance on the link prediction and node classification tasks (see Table~\ref{tab.T_link} and Figure~\ref{fig.node_class} respectively), it struggles on the path prediction task, especially on the Cora\_ML dataset. We believe this is because ONE only explores first-order proximity for structure  preservation, but not higher-order proximity. In contrast, methods with the capability of learning from high-order proximity, such as G2G and TEA, are superior to those that do not, e.g. ONE. }

{Further, we test the path prediction performance (F$_1$ score) of neighbourhoods of different orders (i.e.\ path lengths) on the Citeseer dataset, as shown in Figure~\ref{fig.sp_pp}, where ``no'' on the x-axis means path length is greater than 4 or no path exists between two nodes. It is worth noting that we randomly sample 80\% of training samples as the training set and the remaining for testing. 
	We can observe that some of the methods, such as DW and ONE, suffer a relatively larger performance decrease with the increasing orders of neighbours, as shown in Figure \ref{fig.sp.a}, possibly because the higher-order neighbours require more hierarchical information to be preserved, which these methods do not provide as they cannot learn the discriminative structural patterns for different types of neighbours. 
	In contrast, our adaptive margin component can effectively model the hierarchical information by embedding differentiated shortest distances, and thus \sys does not suffer a large performance decrease on the higher-order neighbours. 
	Moreover, we can observe a moderate performance climb for the ``no'' category for all methods. We conjecture the main reason is that the number of training samples for the ``no'' class is much greater than the  others due to the fact that the class of ``no'' includes all the nodes with their path greater than 4 and those disconnected nodes, resulting in those models being sufficiently trained on this class. 
	Similarly, with respect to the discrete models in Figure~\ref{fig.sp.b}, \sysq also gains the best performance with a 2$\sim$3-point lift compared  with SNEQ~\cite{he2020sneq}, the second best model.
}

%
%

\begin{table*}[htb]
	\centering
	\caption{{Path prediction results on the three attributed networks. The top block shows the results of continuous embeddings while the bottom  shows the results of discrete embedding results. The best results in each group is bold.} }
	\label{tab.path_pred}
	\resizebox{0.9\textwidth}{!}{
	
	\begin{tabular}{llcccccccccccc}
		\toprule
		\multirow{13}{*}{\rotatebox{-90}{{Continuous}} } & \multirow{2}{*}{{Methods}} & \multicolumn{4}{c}{{Citeseer}}   & \multicolumn{4}{c}{ { Cora\_ML }}        & \multicolumn{4}{c}{{DBLP}  }       \\ 
		\cmidrule(lr){3-6}\cmidrule(lr){7-10}\cmidrule(lr){11-14}  
		
		&             & {20\%}   & {40\%}   & {60\% }  & {80\%}      & {20\%}   & {40\%}   & {60\%}   & {80\%}      & {20\%}   & {40\%}   & {60\%}   & {80\%}       \\  \cmidrule{2-14}
		& {DW}       & {blue}31.02 & {blue}31.71 &{blue}32.05 & {blue}32.54  & {blue}28.02  &  {blue}30.11 & {blue}33.25 & {blue}34.34 & {blue}30.13 & {blue}31.57 & {blue}32.28 & {blue}33.47 \\
		& {TADW }     & 34.57 & 36.05 & 36.32 & 37.02  & 31.57  & 32.03  & 35.42 & 36.83 & 29.52 & 33.23 & 34.25 & 35.06 \\
		& {ONE}       & 32.21 & 33.38 & 35.92 & 36.65  & 25.71  & 26.84  & 27.06 & 29.15 & 26.58 & 30.64 & 32.01 & 33.40   \\
		& {SEANO}   & 36.76 & 37.25 & 38.13 & 40.11  & 30.12  & 31.29  & 33.57 & 35.80 & 31.54 & 35.29 & 36.19 & 36.87 \\
		& {TEA}     & 41.10 & 42.94 & 44.16 & 45.53  & 36.63  & 37.94  & 39.13 & 42.64 & 36.53 & 39.89 & 41.32 & 41.48  \\
		
		& {H\_GCN}   & 42.06 & 43.48 & 43.76 & 46.95  & 32.01  & 33.41  & 36.23 & 37.95 & 37.84 & 39.45 & 40.13 & 42.14 \\
		& {G2G}      & 45.73 & 47.26 & 50.39 & 51.42  & 37.72  & 38.59  & 39.06 & 41.81 & 39.36 & 40.28 & 41.60 & \textbf{43.71} \\
		&{ \textbf{d-SNEQ} }&\textbf{49.41} &\textbf{50.43} &\textbf{52.25} & \textbf{53.31}   &\textbf{40.22} &\textbf{41.40} &\textbf{43.52} & \textbf{44.73}   & \textbf{41.21} &\textbf{42.10} & \textbf{42.82} & {43.46}  \\			
		
		\midrule                             
		\multirow{6}{*}{\rotatebox{-90}{{Discrete}} }  
		& {LR}      & 21.11 & 22.10 & 22.57 & 23.01 & 21.71  & 22.21  & 23.13  & 23.64  &  21.16 & 22.01 & 23.57 & 24.31 \\
		& {SH}      & 23.02 & 24.51 & 25.84 & 27.86 & 26.02  & 27.52  & 28.35  & 29.16  &  23.03 & 24.12 & 25.35 & 27.06 \\
		& {DCF}     & 24.34 & 26.27 & 27.53 & 28.54 & 25.15  & 26.86  & 27.97  & 28.01  &  24.52 & 24.91 & 25.23 & 26.54  \\
		& {BANE}    & 30.58 & 32.61 & 33.45 & 34.12 & 27.48  & 28.03  & 29.15  & 31.62  &  25.29 & 27.65 & 28.90 & 29.92  \\
		& {NetHash} & 28.71 & 30.84 & 31.06 & 32.35 & 28.71  & 29.84  & 30.36  & 31.35  &  25.41 & 26.84 & 27.56 & 30.35 \\
		&{ SNEQ$^\#$} & 29.79 & 33.03  & 33.92 & 35.04  & 29.35 & 30.86 & 32.42  & 33.95  & 26.76 & 27.77 & 29.01 & 30.40\\
		&{\textbf{d-SNEQ$^\#$}} & \textbf{32.11}  & \textbf{34.61} & \textbf{35.07} & \textbf{36.42}  & \textbf{30.21} & \textbf{32.31} & \textbf{33.07}  & \textbf{34.12}  & \textbf{28.11} & \textbf{30.61} & \textbf{31.47} & \textbf{32.12}  \\
		
		\bottomrule
	\end{tabular}
	}
\end{table*}

\subsection{Node Recommendation}
In the node recommendation task, given a node, an embedding model ranks all nodes according to a distance measurement and recommends the closest node. 
This task is widely applied in social and e-commerce networks, specially in the recommendation scenario. 
Following the settings of INH-MF~\cite{lian2018high}, $90\%$ of neighbours of each node are used to train each embedding model while the remaining $10\%$ neighbours are reserved for testing. 
We use NDCG@50 as the evaluation metric, and the final results are averaged over $10$ runs. 
Table~\ref{tab.node_recom} shows the results of node recommendation, where we can obtain the following observations: 
\begin{figure}[]
	\centering
	\subfigure[Continuous embeddings.]{
		\includegraphics[width=0.465\linewidth, ]{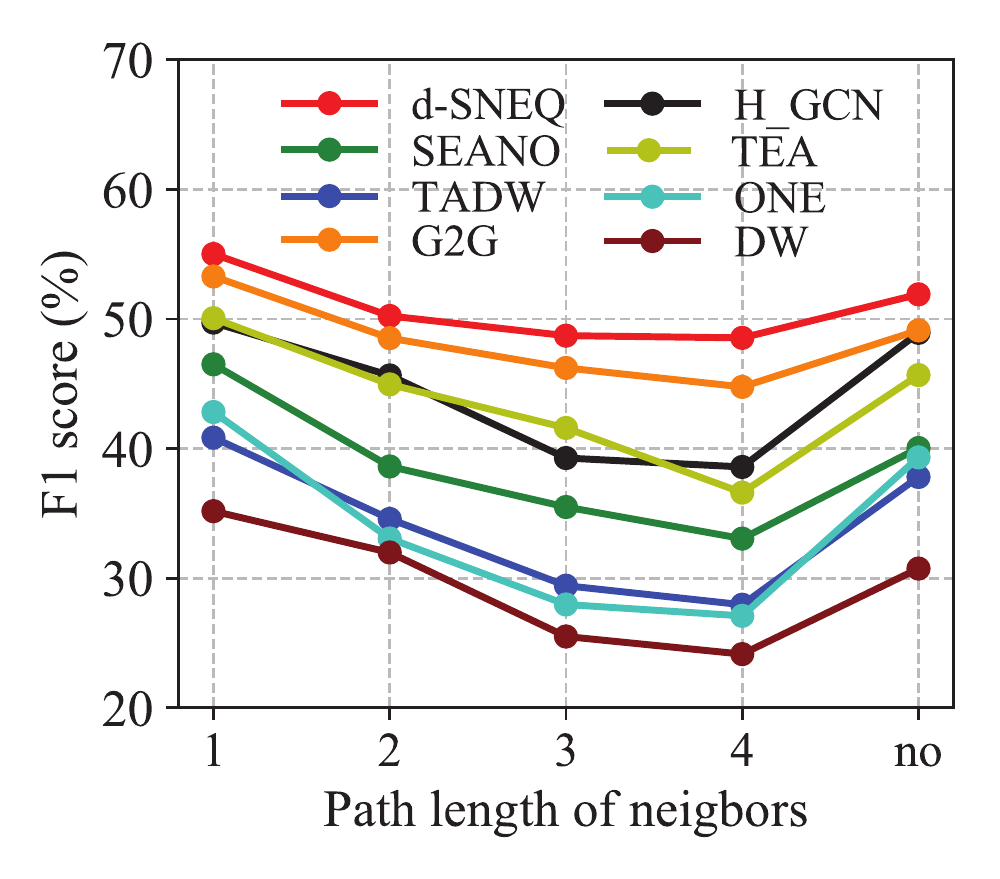}
		\label{fig.sp.a}
	}
	\subfigure[Discrete embeddings.]{
		\includegraphics[width=0.465\linewidth, ]{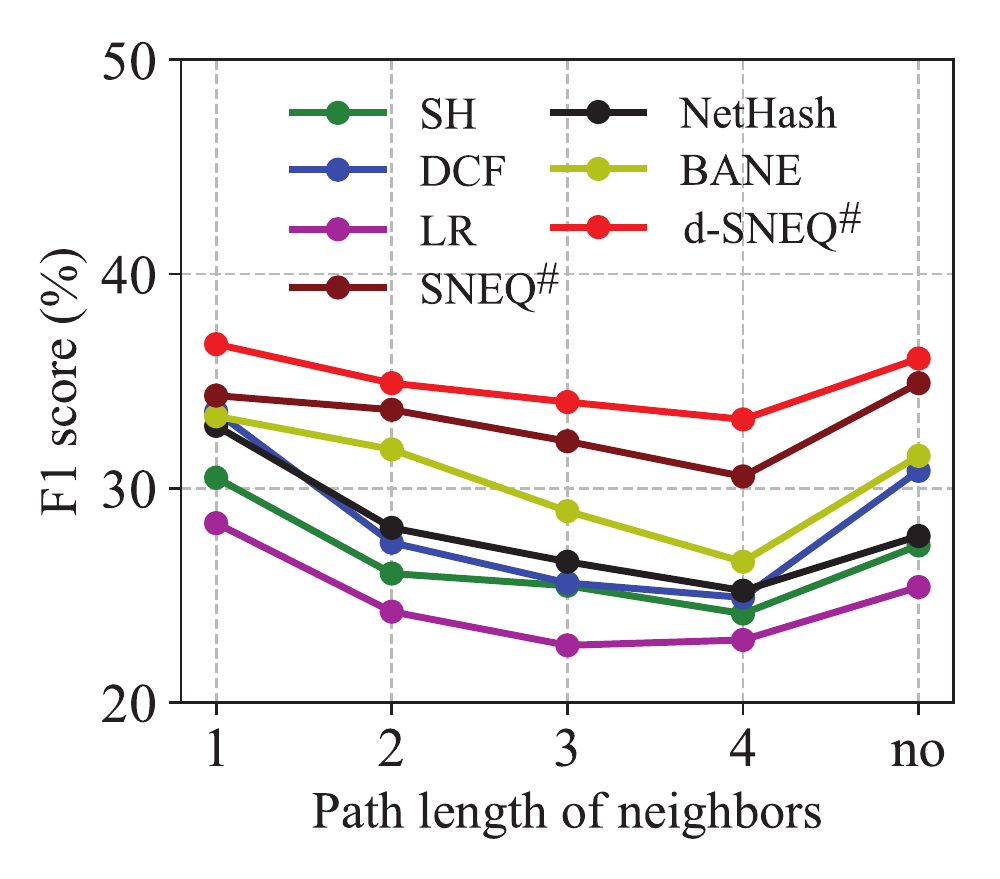}
		\label{fig.sp.b}	
	}
	\caption{ 
		Path prediction F$_1$ scores  of each type of neighbors on Citeseer with 80\% training samples. The x-axis denotes the increasing order  of neighbors from the left to right and   ``no'' means  two nodes' path length is infinite.  }
	\label{fig.sp_pp}
\end{figure}

(1) In terms of the continuous embedding methods, our method outperforms others on all the four datasets. 
Specifically, our method is superior to the most competitive methods H-GCN and G2G by about 2--3 percentage points. 
G2G adopts a rank loss to preserve neighbourhood information, but one difference from ours is that we learn adaptive margins for different neighbours: more distant neighbours will result in a larger margin, i.e., a larger penalty, while closer neighbours are separated by a relatively small margin. 
Hence, our method is more advantageous for the preservation of first-order proximity.

(2) In the second part of Table~\ref{tab.node_recom}, our method shows even larger superiority with an average 4--9 percentage points higher than the best hash method NetHash. 
SH exhibits the worst performance, and the reason is that, as we stated above, hash codes are learned separately in a two-step way. 
It is worth noting that because of excessive memory usage by DCF and BANE, we did not obtain their results on the large network cDBLP. 

\begin{table}[!h]
	\centering
	\caption{Node recommendation results on the four networks, measured by NDCG@50. The upper block shows the results of the continuous embedding methods while the lower block contains results calculated from embeddings reconstructed from discrete codes. Best results in each block are \textbf{bolded}.}
	\label{tab.node_recom}
	\begin{tabular}{l*{5}{c}}
		\toprule
		Method   & Citeseer& Cora\_ML    & DBLP  & cDBLP \\ \midrule
		DW  & 35.13 & 41.52  & 11.68 & 16.49  \\  
		SEANO     & 56.63 & 48.50  & 12.15 & 17.69  \\  
		
		ONE    & 54.02 & 44.27  & 14.13 & N/A     \\ 
		G2G   & 58.31 & 48.12  & 12.37 & 18.53 \\  
		
		TEA  & 58.26 & 47.53  & 12.12 & 19.30  \\  
		H-GCN     & 59.11 & 47.07  & 10.73 & 20.53 \\  
		\sys    & \textbf{61.21} & \textbf{50.41} & \textbf{14.78} & \textbf{23.59}  \\ \midrule
		SH        & 36.04 & 35.31  & 4.16  & 10.92  \\ 
		DCF       & 21.60 & 20.35 & 4.62  & N/A      \\ 
		NetHash       & 41.48 & 38.04  & 4.64  & 12.51  \\  
		BANE       & 35.62 & 37.35 & 8.71  & N/A      \\
		
		SNEQ$^\#$   & {43.41} & {41.82}  & {9.79}  & {19.05}  \\
		\sysq   & \textbf{45.70} & \textbf{43.16}  & \textbf{12.23}  & \textbf{23.51} \\
		\bottomrule
	\end{tabular}
	
\end{table}

\subsection{The Effectiveness of Each Component}

%
%
%

In this section, we will study the effectiveness of each component in our framework. Specifically, we will address the following four questions:

\begin{enumerate}
	\item Does the adaptive margin improve the preservation of structural information compared with a fixed margin? 
	
	\item Is the semi-supervised semantic margin effective for semantic embedding and what is the impact of different values of $T$?
	
	\item Can the  rank loss $\ell_r$ improve the performance of quantisation codes?
	
	\item {Is the differentiable module superior to our previous self-attention-based strategy in SNEQ?}
	
\end{enumerate}

\begin{figure}[!h]
	\centering
	\subfigure[\small{  }]{\label{fig:abl_a}
		\includegraphics[width=0.48\linewidth,  ]{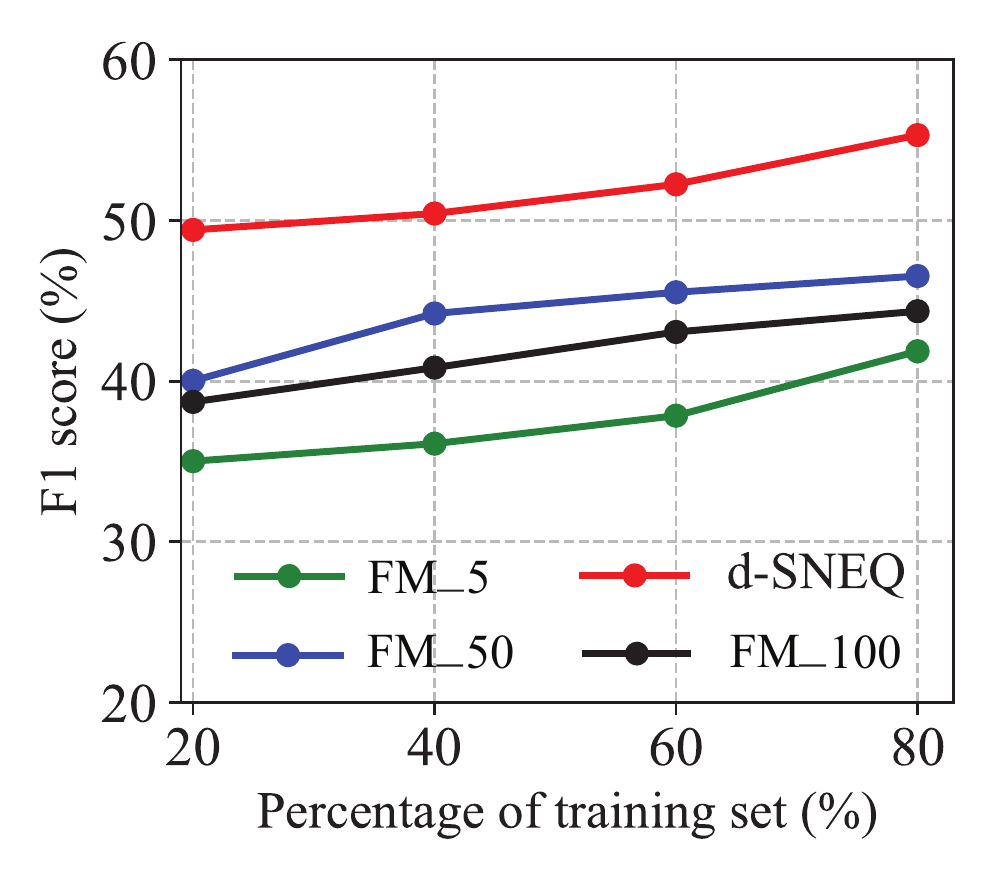}}
	\subfigure[\small{} ]{\label{fig:abl_b}
		\includegraphics[width=0.48\linewidth,  ]{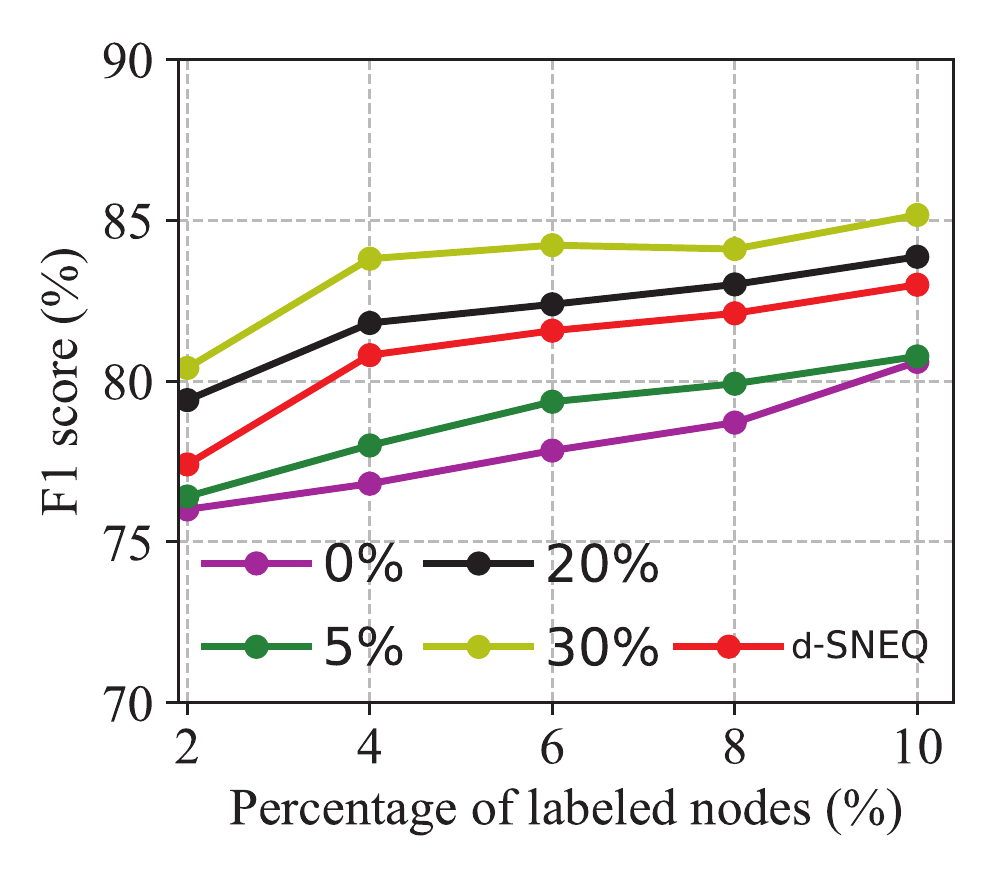}}
	\centering
	\caption{(a) The path prediction results with different fixed margins on Citeseer; (b)The effectiveness of the semantic margin in terms of node classification performance on Citeseer.}
	\label{fig:abl}
\end{figure}

For the first question, we modify the adaptive margin component to a number of different fixed margins: $5$, $50$ and $100$. The reason for the selection of the fixed margins is that we find the fixed margin $50$ shows the best result in our  experiments and the margins close to $50$ do not make a difference and hence we select two faraway margins $5$ and $100$ to represent a small margin and a large margin, respectively.  
Table~\ref{tab.abl_am} shows the link prediction  results on Citeseer, Cora\_ML and cDBLP, where each row with FM denotes a setting of the fixed margin and, Figure~\ref{fig:abl_a} shows the results of path prediction on the Citeseer dataset. Note that experimental settings are kept the same for all margins.

From Table~\ref{tab.abl_am}, it is obvious to see that the adaptive margin outperforms the fixed counterparts. 
In addition, we can observe that too large or too small a margin does not produce competitive results. 
The possible reason is that when the margin is too small, it is hard to separate nodes in the embedding space, but when it is too large, the loss function is more difficult to converge. 
A problem of fixed margins is that a node's neighbours of different hops can be only coarsely divided by the same distance, losing discriminative neighbourhood information in the embeddings. 
In contrast, our adaptive margin can avoid the problem since it can learn different distances for neighbours at different hops, that is, hierarchical structure information. Specifically, those close to the anchor node are separated by a small margin while more distant neighbours by a large margin, which can more precisely model the original neighbourhood information. Figure~\ref{fig:abl_a} further demonstrates this observation, because our adaptive margin gains a higher improvement in terms of the path prediction task, which means that our model can preserve much more high-order or hierarchical structure information compared with different fixed margins.

\begin{table}[ht]
	\caption{Link prediction results with different fixed margins.}
	\label{tab.abl_am}
	\centering
	\begin{tabular}{l*{4}{c}}
		\toprule
		{Margin} &  {Citeseer}                           &  {Cora\_ML}                         & {cDBLP}                    \\  \midrule
		FM=5                     & 88.43                                          & 90.12                                      &                   87.26              \\ 
		FM=50                    & 95.62                                           & 94.03                                     &           91.52                             \\ 
		FM=100                   &  94.37    &  {95.01}   &   90.84\\ 
		Adaptive (full model)                      &  \textbf{96.50}    & \textbf{97.22}   &    \textbf{92.24}                  \\ \bottomrule
	\end{tabular}
\end{table}

The second question is addressed in Figure~\ref{fig:abl_b} and Table~\ref{tab.T_link}. Specifically, Figure~\ref{fig:abl_b} shows the effect of the semantic margin on node classification  on the Citeseer network. 
0\% stands for the model variant removing the semantic loss, i.e., no label information is provided to the model. 
We also evaluate the impact of different amounts of labelled nodes ($T$) respect to all the nodes, i.e., 5\%, 20\% and 30\%, used in the semantic margin loss. 
Noting that our full model has a $T$ value of 10\%. 
Obviously, with label information present, our model outperforms the variant with the supervision signal removed (0\%).
Also, it can be seen that a larger amount of supervision information results in improvements of node classification performance. 

Link prediction results on the three datasets are shown in Table~\ref{tab.T_link}. 
Interestingly, in Table~\ref{tab.T_link} we can observe that as the value of $T$ increases, link prediction performance deteriorates. 
From Figure~\ref{fig:abl_b} we have observed an opposite trend, that with the increase of $T$, node classification performance improves.
This is because as larger $T$ brings more supervised information, it leads to more conflicts between the semantic margin and the adaptive margin, and thus negatively impacts the preservation of neighbourhood structure.
Hence, to balance both tasks, we choose $T$=10\% for our final model. 

\begin{table}[htb]
	\centering
	\caption{
		The impact of different $T$ values on link prediction performance.}
	\label{tab.esn2}
	\begin{tabular}{l*{6}{c}}
		\toprule
		{$T$ value} &  {Citeseer}                 &  {Cora\_ML} &  {cDBLP} \\ \midrule
		5\%                     &  \textbf{96.75}   & \textbf{97.53}  &  \textbf{94.12}     \\
		10\% (full model)              & {96.50}    & {97.22}   &     {92.24}  \\ 
		20\%                    &  93.32   & 94.43  &  90.95     \\  
		30\%                     &  90.56   & 91.59  &  89.59     \\  
		\bottomrule
	\end{tabular}
	\label{tab.T_link}
\end{table}

The third and fourth questions  are addressed by Table~\ref{tab.selfq}, which shows the quantisation results  of  link prediction on the Citesser, Cora\_ML and DBLP datasets, where $-\ell_r$ stands for the model removing the constraint of rank loss in Equation~(\ref{eq.am1}) and +SA means that we replace {Eq.\ref{eq:gumb}}  with our self-attention module in SNEQ~\cite{he2020sneq}.
With the rank loss removed, we can observe a performance dip of approx.\ 1--2 percentage points, which demonstrates $\ell_r$'s role in improving   quantisation quality through equipping the model with structure information.

{
	Similarly, comparing +SA with the full model, a performance dip of about 1 percentage point occurs when we replace the Gumbel-softmax function with self-attention. However, it is worth noting that +SA still outperforms SNEQ$^\#$ by one point, which confirms that our DNN-based quantisation is better than non-deep neural network quantisation in SNEQ~\cite{he2020sneq}.
}

\begin{table}[!h]
	\centering
	\caption{The impact of the rank loss on quantisation codes on three datasets where the percentage of training set is set to $2\%$.}
	\label{tab.selfq}
	\begin{tabular}{l*{5}{c}}
		\toprule
		Method  &  {Cora\_ML} & {Citeseer} & DBLP  & cDBLP    \\ \midrule
		-$\ell_r$ & 91.24   &  92.31   &  93.65  &91.59  \\  
		
		+SA &{92.91}    &  {93.57}   &  {93.82} & {92.46} \\
		Full model & {93.84}    & {94.24}   & {94.71} & {93.54}    \\\bottomrule
	\end{tabular}
\end{table}

\subsection{Space and Time Efficiency }\label{sec:ste}

A main advantage of quantisation is the reduction in storage cost and retrieval time. 
We compare the efficiency of the continuous and discrete embeddings and we choose SH as the representative hash method. 

The space cost results are shown in Table~\ref{tab.storage}, where the first row, {Float}, denotes the storage cost of continuous embeddings with dimension $L = 128$.  
The second row is the result of hash codes generated by SH with the same code length  $64$ bits, same as ours. 
The space and time efficiency results are shown in Table~\ref{tab.storage} and~\ref{tab.time} respectively. 
Empirically, it can be observed in Table~\ref{tab.storage} that, the larger the network, the more reduction in storage quantisation brings: for cDBLP, the original embeddings are approx.\ 50 times larger than ours. 
Compared to hashing methods such as SH, it can be observed that quantisation uses slightly more space as it stores $M$ codebooks.
However, the evidence in the previous three subsections demonstrates hash methods' substantially inferior task performance to quantisation. 
In fact, the additional storage cost of quantisation over hashing is the codebooks, at approx.\ $512$KB for ours. 
Therefore, for larger networks, hashing methods' savings in storage are minuscule and negligible given modern hardware.

Table~\ref{tab.time} shows the average retrieval time of node recommendation in milliseconds. 
For real-valued embeddings, we use the Euclidean distance to measure node similarity, while for quantised codes we use a pre-calculated table to look up their similarity, same as for our method. 
Quantisation achieves up to 70 $\times$ retrieval speedup over the real-valued embeddings. 
Hashing methods, represented by SH, are more time-efficient than our model. 
However, as we noted above, this is achieved with substantially degraded task performance. 

\begin{table}[!h]
	\caption{Storage cost (MB) of different embeddings.}
	\label{tab.storage}
	\centering
	\resizebox{0.7\columnwidth}{!}{
		\begin{tabular}{l*{5}{c}}
			\toprule
			Method  &  {Cora\_ML} & {Citeseer} & DBLP &  cDBLP   \\ \midrule
			Float &  {2.145}    & {1.513}    & 9.562   & 155.31 \\  
			SH &  \textbf{0.031}    &  \textbf{0.026}    &  \textbf{0.153}    & \textbf{2.524} \\ 

			\sysq & 0.097 &0.109 &0.271 & 2.997  \\\bottomrule
		\end{tabular}
	}
\end{table}

\begin{table}[!h]
	\caption{Average retrieval time cost (ms) of different embeddings.}
	\label{tab.time}
	\centering
	\resizebox{0.7\columnwidth}{!}{
		\begin{tabular}{l*{5}{c}}
			\toprule
			Method  & {Cora\_ML} & Citeseer  & DBLP  & cDBLP  \\  \midrule
			Float &  {42.7}   &  {71.2}   & 191.2  & 3,923.5 \\  
			SH & \textbf{0.7}    &  \textbf{1.0}    &  \textbf{2.7}    & \textbf{34.7} \\  

			\sysq & {1.1} &  {1.4} &  {4.4} &  {56.3}\\   \bottomrule
		\end{tabular}
	}
\end{table}

\section{Conclusion}
Network embedding learns fixed- and low-dimensional vector representations for nodes in a network, aiming at preserving node attributes and the structure information in the network. 
With the growth in network sizes, space- and time-efficient embedding becomes an increasingly challenging problem. 
In this paper, we propose \sys, an end-to-end network embedding and quantisation framework. 
\sys is trained in a semi-supervised manner, which simultaneously preserves network structure and semantic information into the embedding space while compressing the embeddings by self-attention-based product quantisation. 
Specifically, we incorporate a adaptive margin loss for preserving network structure information, a semantic margin loss for semantic space learning, and a DNN-based quantisation loss to learn compact codes. 
Moreover, we propose a new evaluation metric, path prediction to better and more directly evaluate a model's ability to preserver higher-order proximity.  
In standard evaluation tasks on four diverse networks, our method outperforms state-of-the-art network embedding methods. 
In addition, due to quantisation, \sys significantly reduces storage footprint and accelerates query time for node recommendation. 
Compared to hashing methods, \sys achieves substantially superior task performance while maintaining comparable space and time efficiency. 

\section*{Acknowledgment}

\bibliographystyle{IEEEtran}
\bibliography{acmart}

\end{document}